\documentclass[11pt,a4paper]{article}
\usepackage[hyperref]{emnlp2020}
\usepackage{times}
\usepackage{latexsym}

\usepackage{microtype}

\aclfinalcopy 
\def\aclpaperid{304} 

\usepackage{amsmath}
\usepackage{multirow}
\usepackage{amssymb}
\usepackage{graphicx}

\title{\textsc{Orchard}: A Benchmark For Measuring Systematic \\ Generalization of Multi-Hierarchical Reasoning}

\author{Bill Tuck Weng Pung*, Alvin Chan \\
    Nanyang Technological University\\
  \texttt{*pung0011@e.ntu.edu.sg} \\}
  
\date{}

\begin{document}
\maketitle

\begin{abstract}
The ability to reason with multiple hierarchical structures is an attractive and desirable property of sequential inductive biases for natural language processing. Do the state-of-the-art Transformers and LSTM architectures implicitly encode for these biases? To answer this, we propose \textsc{Orchard}, a diagnostic dataset for systematically evaluating hierarchical reasoning in state-of-the-art neural sequence models. While there have been prior evaluation frameworks such as ListOps or Logical Inference, our work presents a novel and more natural setting where our models learn to reason with multiple explicit hierarchical structures instead of only one, i.e., requiring the ability to do both long-term sequence memorizing, relational reasoning while reasoning with hierarchical structure. Consequently, backed by a set of rigorous experiments, we show that (1) Transformer and LSTM models surprisingly fail in systematic generalization, and (2) with increased references between hierarchies, Transformer performs no better than random.
\end{abstract}

\section{Introduction}


Sequential models like the Long-Short Term Memory (LSTM) \cite{hochreiter1997long} and Transformer \citep{vaswani2017attention} have achieved state-of-the-art (SOTA) performance on a myriad of NLP tasks, such as language modeling \citep{melis2019mogrifier}, machine translation \citep{edunov2018understanding}, summarization \cite{yan2020prophetnet} and document classification \citep{adhikari2019rethinking}. Language is hierarchical in nature, where documents are composed of sentences, and in turn composed of words. These hierarchical structures are intricately cross-referencing, with words referencing others in different sentences or even paragraphs. For example, if we use the term 'aforementioned SOTA models' in this sentence, the reader needs to relate it to the terms 'LSTM' and 'Transformer' in the first sentence of this paragraph. In our paper, we ask this central question: Do these SOTA models exploit the ubiquitous relational hierarchies in NLP? Our experiments show they do not, and surprisingly fail to reason with multiple hierarchies when tested on systematic generalization of our proposed task. Transformer does no better than random chance with larger hierarchies.

\begin{figure}[ht]
    \centering
    \includegraphics[width=\columnwidth]{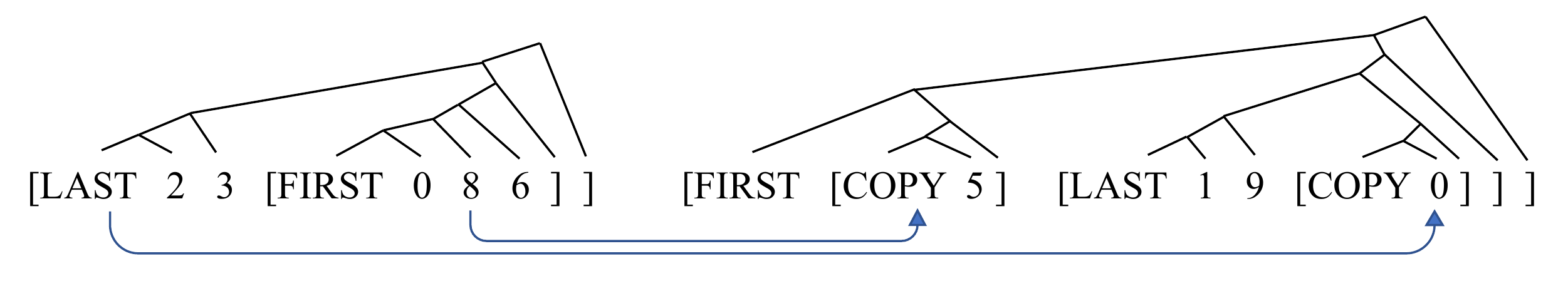}
    \caption{Example parse of an \textsc{Orchard} sequence showing the ground truth parse tree. Correct output is 0, 8. Blue arrows indicate the referencing done by COPY operators in the second tree}
    \label{fig:tree}
\end{figure}

We study these models' ability of capturing hierarchical structure without explicit parse trees, by evaluating the bi-directional LSTM with attention and the Transformer models on a novel multi-hierarchical reasoning task. To mitigate effects of spurious correlations and annotation artifacts in natural language datasets \citep{niven2019probing}, we propose Operating Relational-Cum-Hierarchical Arithmetic Reasoning Dataset (\textsc{Orchard}), a dataset comprising sequences of numbers and nested mathematical operators. The task is as follows: Given two trees with values 0-9 and operators \{FIRST, LAST, MIN, MAX, COPY\}, evaluate both trees at their roots. An example is illustrated in Figure \ref{fig:tree}.

\subsection{Contributions}
\textbf{Diagnostic dataset} We design a sequence-to-sequence dataset\footnote{\url{https://github.com/billptw/Orchard}} for measuring relational reasoning on hierarchical sequences. We evaluate six variants of the \textsc{Orchard} tasks, where models resolve the values of increasingly nested mathematical operations of different types, with explicit parses in the form of parenthesized lists.

\textbf{Model analysis and generalization tests} We perform experimental evaluation on Transformer and attentional Bi-LSTM architectures, and show that these models perform surprisingly badly on systematic generalization, i.e. when trained on a train set of depths 3 to 6, and tested on separate bins of depth 3 to 12, the model performance degrade significantly beyond depth 7. We show that the LSTM model is better able to systematically generalize to larger depths than the Transformer.

\section{Related Work}
Learning to induce hierarchical structures from sequential data has shown tremendous potential in many recent works \cite{wang-etal-2019-tree,NIPS2019_8748,shen2018ordered,shen2017neural,yogatama2016learning,choi2018learning,drozdov2019unsupervised,jacob2018learning}. After all, many forms of sequential data, especially language, are intrinsically hierarchical in nature.




ListOps \cite{nangia2018listops}, subejct-verb agreement \citep{linzen2016assessing}, and logical inference \cite{bowman2015tree}, were designed to test hierarchical reasoning in tree induction models. \citet{tran2018importance} evaluate Transformer and LSTM on the latter two tasks, suggesting the importance of recurrence in modeling hierarchical structure. These studies evaluate models on tasks requiring reasoning about single parse trees at a time. Our proposed work examines relational reasoning with multiple hierarchies, and is the first to do so (to the best of our knowledge).

\section{\textsc{Orchard} Dataset}

\subsection{Description}
Sentences in the \textsc{Orchard} dataset comprises nested mathematical operations on lists of single-digit integers, written in polish notation for easy parsing into syntax trees. Each sentence has a corresponding single-digit integer solution. For example, [FIRST [LAST 7 3 ] 2 0 9 ] has the solution 3. Each operation has starts with an opening square bracket and ends with a corresponding closing bracket, with the encapsulating constituents making up its operands. In the previous example, LAST operates on \{7,3\} while FIRST operates on \{3,2,0,9\}. 

An input sequence comprises two sentences delimited by the character 'X'. The first sentence contains only positional operators \{FIRST, LAST\}, or comparative operators \{MIN, MAX\}. The second sentence is similar in composition, but also has the relational operator \{COPY\}, which references values in the first sentence. We define the operators in further detail below:

\begin{itemize}
    \item FIRST: the first item of a given list. The sentence [FIRST 2 6 0 1 ] corresponds to 2.
    \item LAST: the last item of a given list. The sentence [LAST 2 6 0 1 ] corresponds to 1.
    \item MIN: the smallest item of a given list. The sentence [MIN 2 6 0 1 ] corresponds to 0.
    \item MAX: the largest item of a given list. The sentence [MAX 2 6 0 1 ] corresponds to 6.
    \item COPY: the $n^{th}$ of a given tree in level order. The sequence [MAX 2 6 0 1 ] X [COPY 1 ] corresponds to 6, 2. The COPY operator only has one argument in its list.
\end{itemize}

The \{FIRST, LAST\} operators require the model to select from a small range of values within the list, purely relying on the position of the item in the list. This makes it relatively easier than the comparative operators \{MIN, MAX\}, where the model has to compare all items within the list, thus having to store the values of the list in memory.

The COPY operator requires the model to store a level order tree traversal of the first sequence in memory, and evaluate the node at the specified position. This makes it a much harder task, as the model has to accurately identify the correct parse of the first tree in order to reliably evaluate the COPY value.


\subsection{Generation}
To generate the \textsc{Orchard} dataset, we assign to the root node of the binary tree an operator chosen with uniform probability from the set of available operators for each experiment. Only \{FIRST, LAST, MIN, MAX\} operators (termed branching operators) can have children, with value nodes and COPY nodes being terminal nodes. 

In creating a dataset that better corresponds to the variety of branching structures found in natural language, we need to introduce sparsity through random assignment of branching and non-branching nodes in all depths of the parse tree. Hence, in the generation of the tree, the left and right children of each operator node have a equal probability of being assigned a branching operator node or a terminal node. Terminal nodes are nodes with a COPY operator, or with only integer values.


\section{Experiments}
\subsection{Tasks}
To examine the effects of hierarchical reasoning versus relational reasoning, we run three difficulty variants on two permutations of operators used. We generate a train, validation and test set of 500k, 50k and 50k sequences respectively. The train and validation set comprises depths of tree from 3 to 6 in equal proportions, and test on bins of depths 3 to 12 individually for generalization.

\textbf{Difficulty} To vary the difficulty of the relational reasoning required by the models, we vary the number of COPY operators in the second sentence of each input sequence in the dataset. When generating a sequence, each terminal node has a probability \textit{c} of being assigned the COPY operator, or otherwise be assigned one or two integer values ranging from 0 to 9. We vary $\textit{c}$ in the second tree for each difficulty variant of the task, choosing from $c \in \{0,0.5,1\}$ to generate \textsc{Orchard}-easy, \textsc{Orchard}-medium and \textsc{Orchard}-hard tasks respectively. Hence the second tree an each sequence of the \textsc{Orchard}-hard dataset references the first tree fully, containing no integer values in the raw parse.

\textbf{Operators Used} The purpose of the \textsc{Orchard} task is to test the model's ability to reason hierarchically, and not its ability to approximate mathematical operations. Hence, we minimized the number and type of operations used per task. For each of the three \textsc{Orchard} difficulty levels, we trained models on two variants of the \textsc{Orchard} dataset. The FIRST-LAST variant requires only positional and relational operators \{FIRST, LAST, COPY\}, while the MIN-MAX variant requires only comparative and relational operators \{MIN, MAX, COPY\}. 

\subsection{Experimental Details}
\textbf{Models Examined} We train both Transformers of 44.3M parameters and bi-directional attentional LSTM of 6.2M parameters on the tasks, using the Fairseq\footnote{\url{https://github.com/pytorch/fairseq}} framework to conduct the experiments. We minimize the sum of log probabilities of the correct character via the Adam optimizer \citep{kingma2014adam}. For both models, we use a batch size of 128 split across 4 NVIDIA RTX 2080 GPUs, trained for 500 epochs using floating-point 16 precision. This results in an average training time of 18 hours. We then generate the output classification using beam search evaluated using the model checkpoint with the best validation accuracy, saved every 50 epochs. We include more hyper-parameter details in the appendix.


\begin{figure}[ht]
    \centering
    \resizebox{\columnwidth}{!}{
    \includegraphics[]{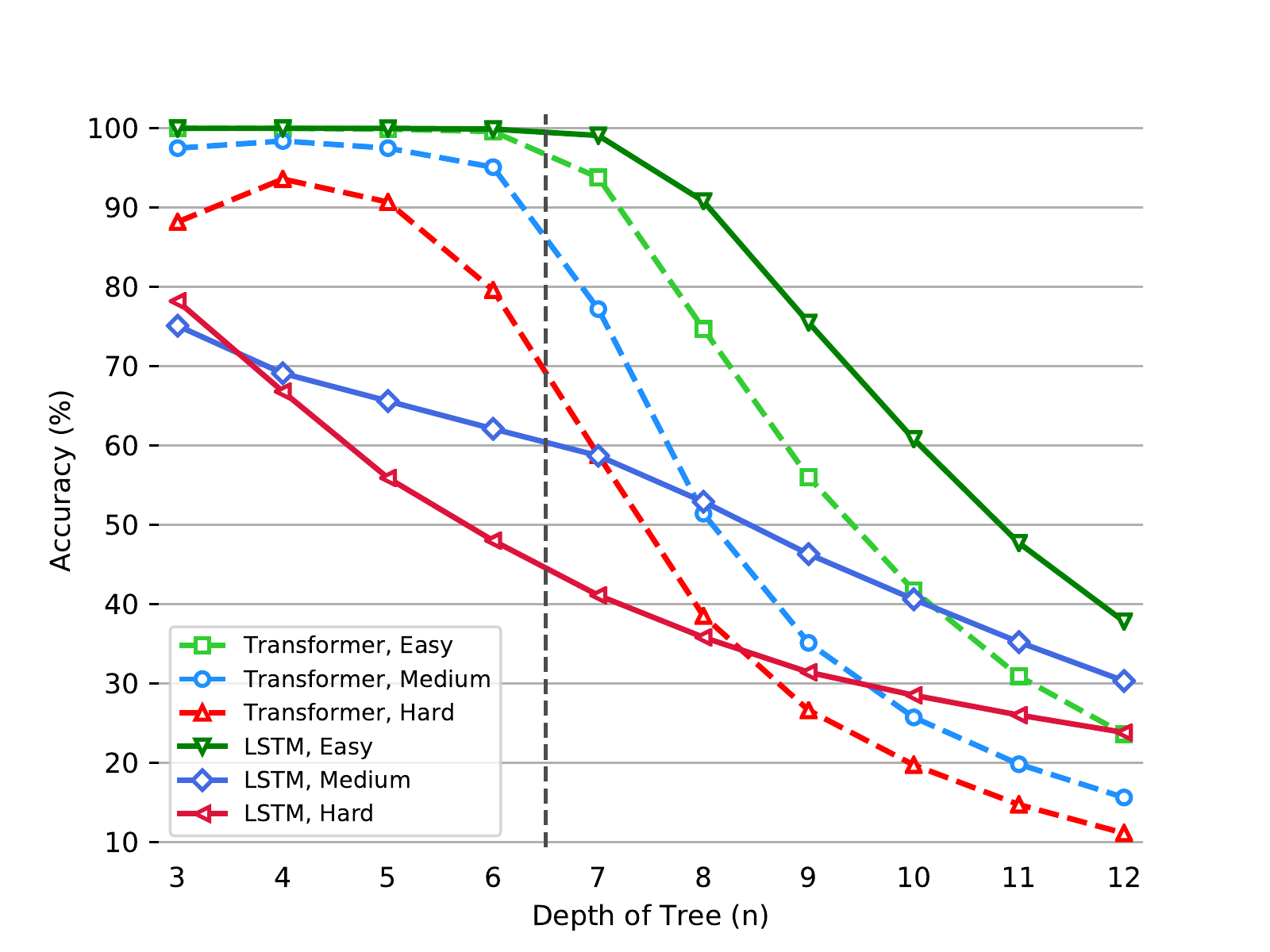}
    }
    \caption{Percentage accuracy for classification on both trees in \textsc{Orchard} tasks comprising only MIN-MAX task. Dashed lines for Transformer model.}
    \label{fig:mm}
\end{figure}

\begin{figure}[ht]
    \centering
    \resizebox{\columnwidth}{!}{
    \includegraphics[]{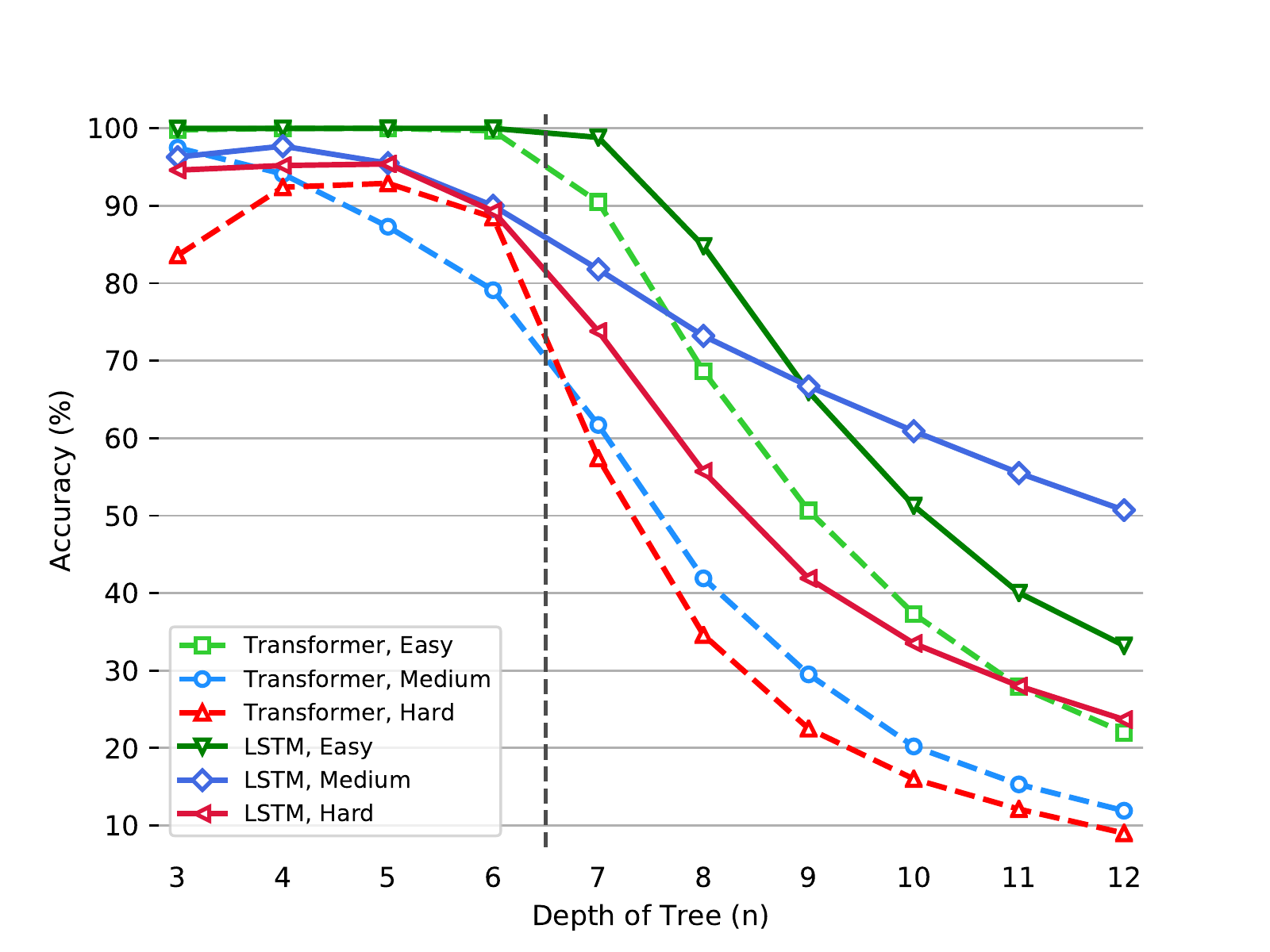}
    }
    \caption{Percentage accuracy for classification on both trees in \textsc{Orchard} tasks comprising only FIRST-LAST task. Dashed lines for Transformer model.}
    \label{fig:fl}
\end{figure}





\begin{figure}[ht]
    \centering
    \resizebox{\columnwidth}{!}{
    \includegraphics[]{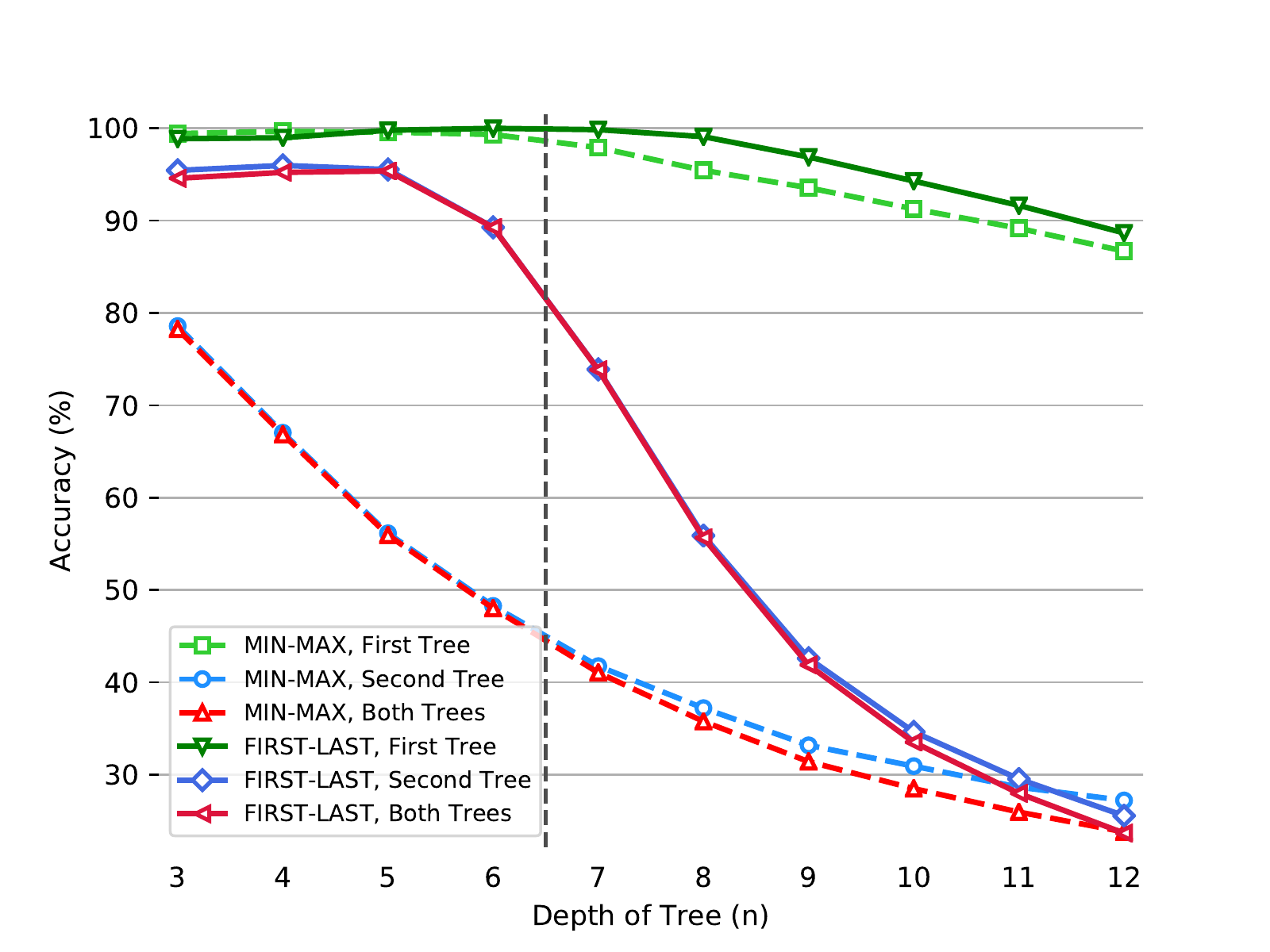}
    }
    \caption{Percentage accuracy for classification of First, Second or Both trees on \textsc{Orchard}-hard task by LSTM model. Dashed lines for MIN-MAX task.}
    \label{fig:hard-lstm}
\end{figure}

\begin{figure}[ht]
    \centering
    \resizebox{\columnwidth}{!}{
    \includegraphics[]{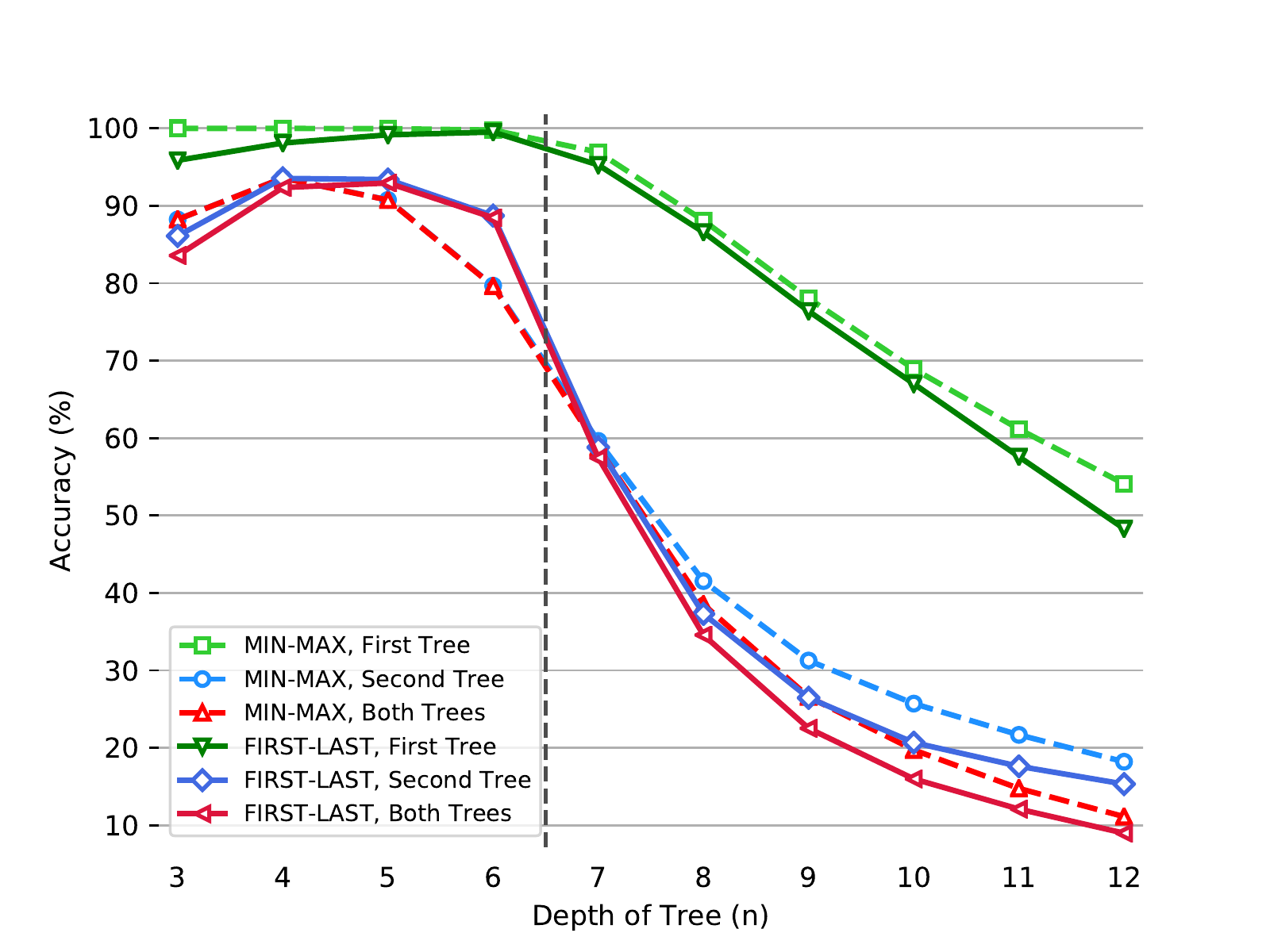}
    }
    \caption{Percentage accuracy for classification of First, Second or Both trees on \textsc{Orchard}-hard task by Transformer model. Dashed lines for MIN-MAX task.}
    \label{fig:hard-tfmr}
\end{figure}

\section{Analysis}

\textbf{\textsc{Orchard} Variants}
From figures \ref{fig:mm} and \ref{fig:fl}, we empirically verify the increased difficulty of ranging the amount of cross-tree referencing through the COPY operator. This is shown by the drop in performance from \textsc{Orchard} easy, medium to hard for both MIN-MAX and FIRST-LAST experiments, on both LSTM and Transformer models. Comparing the comparative MIN-MAX operators with the positional FIRTS-LAST task, we see that both models perform better on the FIRST-LAST dataset for \textsc{Orchard}-med and \textsc{Orchard}-hard. This suggests that reasoning with position within the level of the tree is relatively easy. 


\textbf{Depths of Tree} As the depth of tree N increases, the performance of both models greatly decreases. From figures \ref{fig:mm} and \ref{fig:fl}, we see that both models perform well for depths of tree within the training set (i.e. 3-6). Overall, we see that both models are unable to generalize well. From figures \ref{fig:hard-lstm} and \ref{fig:hard-tfmr}, we see that for the \textsc{Orchard}-hard task, the models can generalize well for the first tree classification. However, the classification score for both trees are held back by poor performance in classifying the second tree, highlighting the difficulty of hierarchical reasoning required in evaluating the COPY operator. These results suggest that the models are unable to relationally reason with hierarchical structures, as the task should be trivial to solve otherwise. Without a successful hierarchical parsing strategy, the difficulty of the \textsc{Orchard} task increases exponentially as the depth of tree increases, as the memory required to store the values of long sequences exceed the hidden state size if the model does not learn to resolve sub-trees in correct hierarchical order.

\textbf{Transformer vs LSTM} The Transformer model outperforms LSTM for \textsc{Orchard}-med and \textsc{Orchard}-hard for depths of tree within the training set (i.e. 3-6). However, the LSTM far outperforms Transformers when generalizing for greater depths of tree ($\geq 9$).From figures \ref{fig:hard-lstm} and \ref{fig:hard-tfmr}, we see that LSTM generalizes very well when evaluating the value of the first tree at depth 12, obtaining an accuracy of $86.7\%$ and $888.7\%$, whereas the Transformer model scored $54.1\%$ and $48.3\%$, on MIN-MAX and FIRST-LAST \textsc{Orchard}-hard tasks respectively. At depth of tree 12, the performance by the Transformer model degrades to random chance at $10\%$. These results suggest that LSTM is able to successfully reason hierarchically, generalizing well when parsing the first tree. However, both models are unable to generalize when doing relational reasoning, which is required when evaluating the COPY operator to successfully classify the second tree, leading to poor performance on the overall task.


\section{Conclusion}
Natural language is structured hierarchically with multiple references between hierarchies, where words from one sentence recursively refer to words or phrases in other sentences. How do SOTA models fare on the natural setting of reasoning between multiple hierarchies in an input sequence? To answer this, we introduce \textsc{Orchard}, a diagnostic dataset involving reasoning with multiple hierarchical structures. We empirically show that LSTM and Transformer models are unable to generalize on this setting, despite a small vocabulary of 10 numbers and 3 operators of the task, as the multi-hierarchical inductive bias is not implicitly captured.

\bibliographystyle{acl_natbib}
\bibliography{anthology,emnlp2020}

\end{document}